\documentclass[letterpaper, 10 pt, conference]{ieeeconf}  

\IEEEoverridecommandlockouts                              

\usepackage{graphicx} 
\usepackage{amsmath} 

\usepackage{gensymb} 
\usepackage{booktabs} 
\usepackage[table,xcdraw]{xcolor}

\title{\LARGE \bf
	AutOTranS: an Autonomous Open World Transportation System
}

\author{Brayan S. Zapata-Impata$^{1}$, Vikrant Shah$^{2}$, Hanumant Singh$^{2}$ and Robert Platt$^{3}$ 
	\thanks{*Work funded by the Spanish Ministry of Economy, Industry and Competitiveness (predoctoral grant BES-2016-078290).}
	\thanks{$^{1}$Dept. of Physics, System Engineering and Signal Theory, University of Alicante, Alicante, Spain
		(email: {\tt\small brayan.impata@ua.es})}%
	\thanks{$^{2}$Dept. of Electrical and Computer Engineering, Northeastern University, Boston, Massachusetts, USA
		(email: {\tt\small shah.vi@husky.neu.edu, ha.singh@northeastern.edu})}%
	\thanks{$^{3}$College of Computer and Information Science, Northeastern University,
		Boston, Massachusetts, USA
		(email: {\tt\small rplatt@ccs.neu.edu})}%
}

\begin{document}
	
	\maketitle
	\thispagestyle{empty}
	\pagestyle{empty}
	
	
	\begin{abstract}
		
		Tasks in outdoor open world environments are now ripe for automation with mobile manipulators. The dynamic, unstructured and unknown environments associated with such tasks -- a prime example would be collecting roadside trash -- makes them particularly challenging. In this paper we present an approach to solving the problem of picking up, transporting, and dropping off novel objects outdoors. Our solution integrates a navigation system, a grasp detection and planning system, and a custom task planner. We perform experiments that demonstrate that the system can be used to transport a wide class of novel objects (trash bags, general garbage, gardening tools and fruits) in unstructured settings outdoors with a relatively high end-to-end success rate of 85\%. See it at work at: \verb|https://youtu.be/93nWXhaGEWA|
		
		
		
	\end{abstract}
	
	
	\section{INTRODUCTION}
	\label{sec:introduction}
	
	
	In many cities, trash bags often accumulate through the week, waiting to be picked up. Farm workers must often pick up and carry heavy tools daily. Construction workers spend a lot of time transporting materials through the construction site. These are all labor intensive tasks that could benefit from mobile robotic manipulation. The research into these robots has been growing \cite{Bostelman2015} but there are still significant challenges handling the uncertain character of many outdoor operating environments. Some recent work does utilize mobile manipulators in outdoors scenarios: using pedestrian cross-walks and traffic lights \cite{Chand2012}, moving around a campus to get you a coffee \cite{Pratkanis2013} or working throughout a solar plant \cite{Maurtua2016}. However, none of these systems can handle outdoor pick and place tasks involving novel objects. This paper explores what can be accomplished in this regard by integrating the latest grasping and navigation methods and software.
	
\begin{figure}[t]
		\centering
		\includegraphics[width = 0.375\textwidth, clip = true, trim = 0 34 0 40]{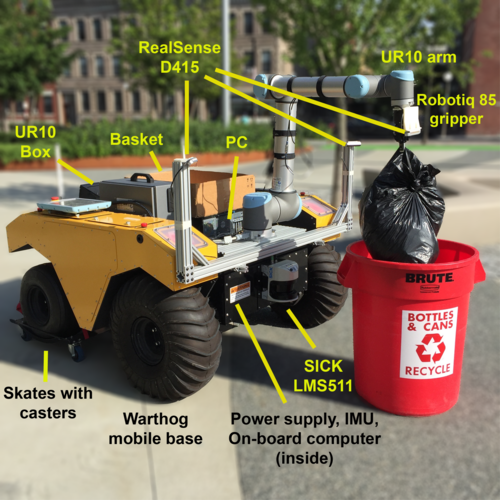}
		\caption{Mobile manipulator used for open world transportation. It comprises a mobile base (Warthog), a robotic arm (UR10) with a gripper (Robotiq 85), a set of cameras (Intel RealSense D415) and a laser sensor (SICK LMS511).}
		\label{fig:golden-fig}
	\end{figure}

	
	We focus on the problem of picking and dropping novel objects in an open world environment. The only input to our system are the the pick and drop points selected by an operator using a map of a previously explored area. Once these points are identified, the robot navigates to the pick location, picks up whatever is found there, transports it and drops it to a bin at the drop location. Our main contributions are as follows. First, we describe a method for navigating to these points autonomously. We propose two transport strategies to solve this task: \textit{collect all} and \textit{collect one by one}. Second, we describe a method for selecting grasps in order to carry out the picking that does not make any assumptions about the objects. Finally, we describe the outdoor mobile manipulator used in detail (Fig.~\ref{fig:golden-fig}). We experimentally characterize the navigation and grasping systems and report success rates and times over four transport tasks on different objects (trash bags, general garbage, gardening tools and fruits).
	
	
	
	
	
	
	\section{RELATED WORK}
	\label{sec:related-work}
	
	Autonomous indoor and outdoor transport has been studied for a long time. In \cite{Prats2008} the authors presented a system for picking books from shelves and navigating a library. Although the system proved to work in this environment, it could only pick books. In \cite{Teller2010} the authors presented a forklift that delivered pallets. However, this system could not handle a more general class of objects. More recently, in \cite{Wang2017}, the authors performed biochemical sampling tasks using a tracked mobile robot equipped with an arm, a gripper, various instruments and visual sensors. However, this system required a remote human operator to teleoperate the system.
	

	
	Planetary explorers face related problems. In \cite{Lehner2018} the authors developed a transportation system that involved picking up an instrument, placing it at some point in the world and later on re-collecting it. Although the process was executed with no human intervention, the environment was modified by using visual tags so the robot could recognize the tools.
	
	A more specific open world transportation task is trash collection. Some early attempts to solve this problem were the OSR-01 \cite{Fuchikawa2005} and the OSR-02 \cite{Nishida2006}. These robots used computer vision to detect and approach loose bottles, then an approximation of the pose of the object  was calculated to find grasps. However, this system could not grasp anything besides bottles and it did not do anything with them after picking them up. Recently, a solution for picking trash on the grass has been proposed \cite{Bai2018}. This system used deep learning for segmenting the grass and detecting objects. It also tracked the object it chose to pick while avoiding the rest of obstacles. Nevertheless, the robot was limited to working on grass and with a set of known objects. Moreover, the robot could not do anything with them after collection.
	
	Grasp detection is a critical part of this system because it enables us to handle completely novel objects. Here, we use GPD, a publicly available grasp detection package~\cite{Pas2017,highprecision}. However, there are a number of other grasp detection methods that would also be relevant here. Perhaps the most well known is the work of~\cite{levine2016learning} who learn closed loop grasping policies from relatively large amounts of real robotic experience. However, it would be challenging to apply this work directly to our outdoor scenario because that system was tuned to work in a specific indoor bin-picking environment. Another important touchpoint is the work of \cite{kopicki2016one} who developed an approach to transferring grasps from a canonical set of model objects to novel objects. However, this method was reported to take a long time to detect grasps (up to 30 seconds) for unsegmented scenes. A faster approach was proposed in \cite{Zapata-Impata2017}, in which the authors defined a set of rules for finding grasps on novel objects. Nevertheless, they assumed medium levels of occlusion, which could not be guaranteed in our open world setting. Most recently, \cite{mahler2017dex} developed a grasp detection system tuned for bin picking. This system achieves success rates comparable to ours, but is specifically tuned for the bin-picking environment.
	
	
	This paper describes a solution to the outdoor autonomous novel object transport task that overcomes some of the limitations of previous systems. Our system only requires as input the approximate pick and drop points in order to carry out the task: it navigates autonomously to the pick point, grasps whatever objects are found there, and then travels to the drop point, where it drops them off into a bin.
	
	
	\section{ROBOT HARDWARE}
	\label{sec:robot-hardware}
	
	The mobile base is the Clearpath Warthog, which offers a payload of 276kg and measures 1.52 x 1.38 x 0.83 (m). Since the default setup gave us steering issues, we disengaged the rear wheels and put them on casters, converting it to a differential drive system. Although this increased our turning radius by 0.43m and added some non-linearity to the steering dynamics, it enabled us to operate the system autonomously.
	
	The manipulator is the Universal Robots UR10, which has 6 Degrees of Freedom (DoF) and a payload of 10kg. The end effector is a Robotiq 2-Finger 85 gripper, with a payload of 5kg and a maximum aperture of 85mm. The arm is mounted at the front of the Warthog with sufficient space around it to rotate without collisions, so it picks up objects from the floor and from a basket on top of the Warthog. This basket (51.0 x 60.5 x 19.5 (cm)) was used to hold a collection of grasped objects so that they could be transported to the drop location. We also mounted on top the UR10 control box and a PC, which runs the higher level processing of the system.
	
	
	For perception, we use three Intel RealSense D415 depth cameras. Two of them are fixed to the two sides in front of the robot pointing downwards to cover the target picking area (see Fig.~\ref{fig:golden-fig}). The third one is mounted on the gripper configured as a hand-eye camera. These cameras generate point clouds from depth by combing a structured light sensor with stereo vision, allowing them to work outdoors, even in moderately bright sunlight. The primary sensor used for vehicle localization is a front mounted single line SICK LMS511 lidar which has a field of view of $190 \degree$.
	
	
	
	
	\section{SYSTEM ARCHITECTURE}
	\label{sec:system-architecture}
	
	\begin{figure}[b]
		\centering
		\includegraphics[width = 0.405\textwidth, clip = true, trim = 0 10 0 0]{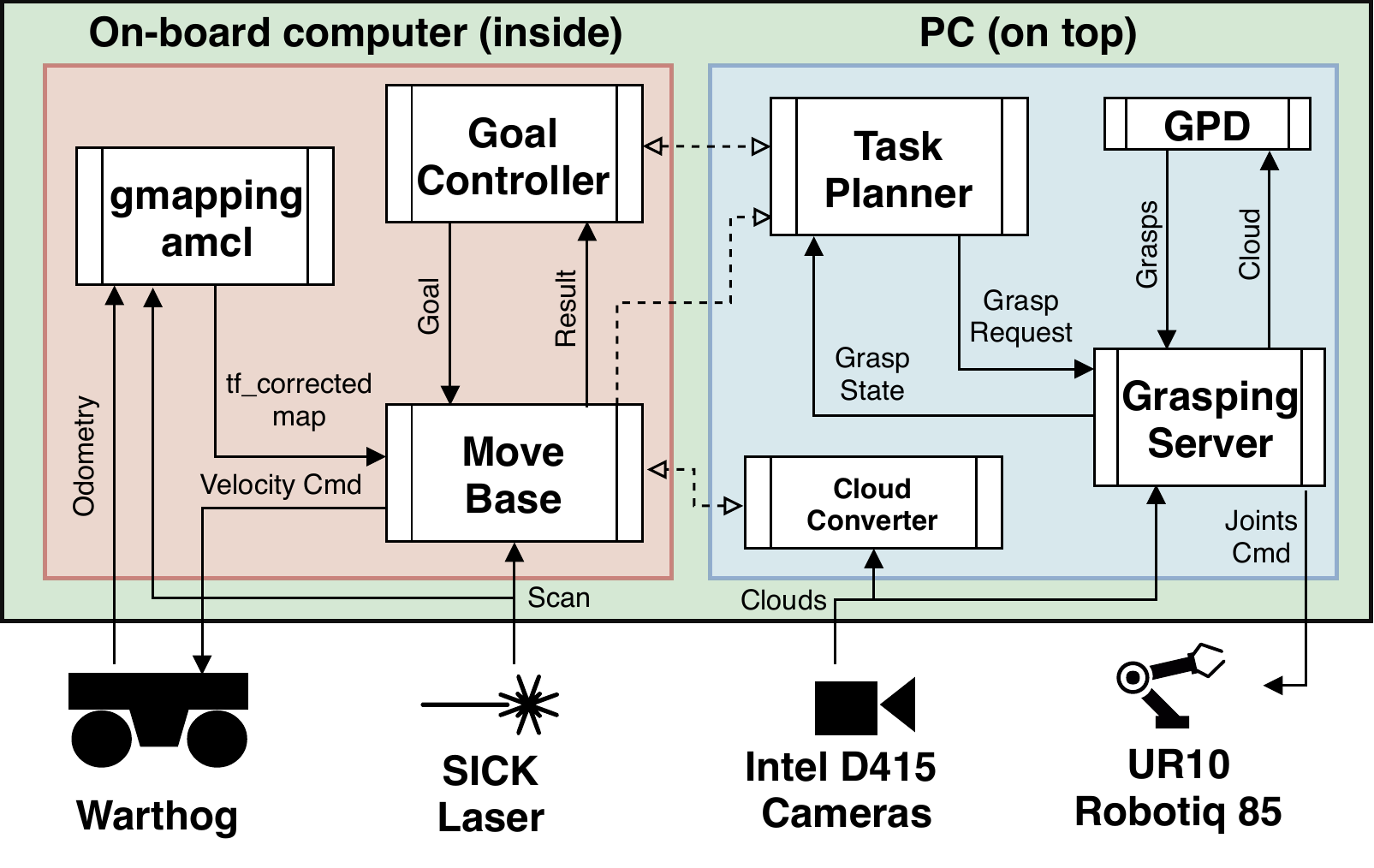}
		\caption{Principal interactions and components of the implemented architecture. Dotted lines are shared ROS messages between \textit{roscores}.}
		\label{fig:proj-nodes}
	\end{figure}
	
	Our system is developed using the Robot Operating System (ROS). The three main parts are: the navigation stack, the grasping stack and the task planner (Fig.~\ref{fig:proj-nodes}). Given their computational requirements, we split the system onto two computers: 1) the on-board PC in the Warthog, which runs all of the navigation stack and 2) the PC mounted on top, which runs the grasping nodes and the task planner.
	
	\subsection{Navigation Stack}
	\label{subsec:nagivation}
	
	\begin{figure}[b]
		\centering
		\includegraphics[width = 0.368\textwidth]{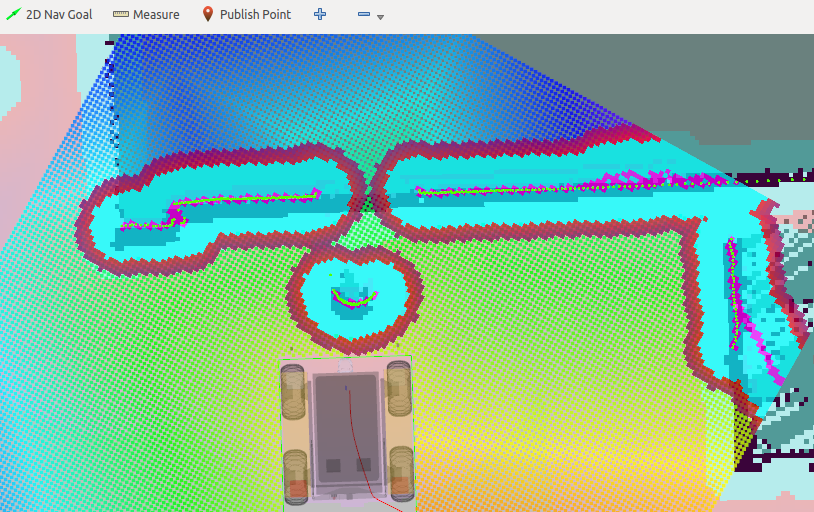}
		\caption{RViz visualization of the robot self-localizing on part of the generated map after configuring the ROS navigation stack.}
		\label{fig:navigation-rviz}
	\end{figure}
	
	The main goal of the navigation is to deliver the robot base to a position such that the target objects are within the manipulator workspace. We use the existing ROS navigation stack, which uses GMapping SLAM for creating maps, AMCL for localization in existing maps and the \textit{move\_base} stack for route planning and control of the robot. GMapping is an implementation of Rao-Blackwellized particle filter to learn occupancy grid maps using raw odometry and laser data \cite{Grisetti2007}. AMCL implements an adaptive Montecarlo localization algorithm which uses an existing map, odometry and laser scans to calculate pose estimates \cite{Fox2002}. The \textit{move\_base} stack implements 2D costmaps, Dijkstra's algorithm based global planner, and a trajectory roll-out local planner, which sends velocity commands to the mobile base \cite{Marder-Eppstein2016}. Given that our lidar is a single line scanner, we assume that our environment has little variation in topography so that it can be navigated using a 2D assumption. After configuring GMapping and AMCL our system was able to generate reliable maps and self-localize accurately (see Fig.~\ref{fig:navigation-rviz}). However, configuring the parameters of the \textit{move\_base} stack proved to be challenging because of the non-linearities added by the casters. This was overcome by increasing the controller frequency to 40Hz.
	
	
	Another set of difficulties arose from our navigation requirements. We wanted to stay far from obstacles when navigating but still get close to the target to work with the objects. To deal with this, we turned off \textit{heading\_scoring} and set the local trajectory scoring parameters \textit{pdist\_scale} (distance to path) greater than \textit{gdist\_scale} (distance to goal), forcing the local planner to stay close to the global plan. Additionally, we set the obstacle inflation in the local cost map (0.5m) smaller than in the global cost map (2.0m). The resulting system finds paths far from obstacles, while still approaches the target objects without triggering collisions.
	
	
	An additional issue we faced was that the lidar could not see short objects, which was required to navigate around objects and to adjust the robot pose precisely for picking. This was addressed using a node called \textit{cloud converter} which reads the point clouds from the two fixed cameras and transforms them into laser scans using the \textit{pointcloud\_to\_laserscan} ROS package. These two scans are then synchronized to the on-board \textit{roscore} to be used with the navigation.
	
	\begin{figure}[b]
		\centering
		\includegraphics[width = 0.40\textwidth, clip = true, trim = 0 0 90 7]{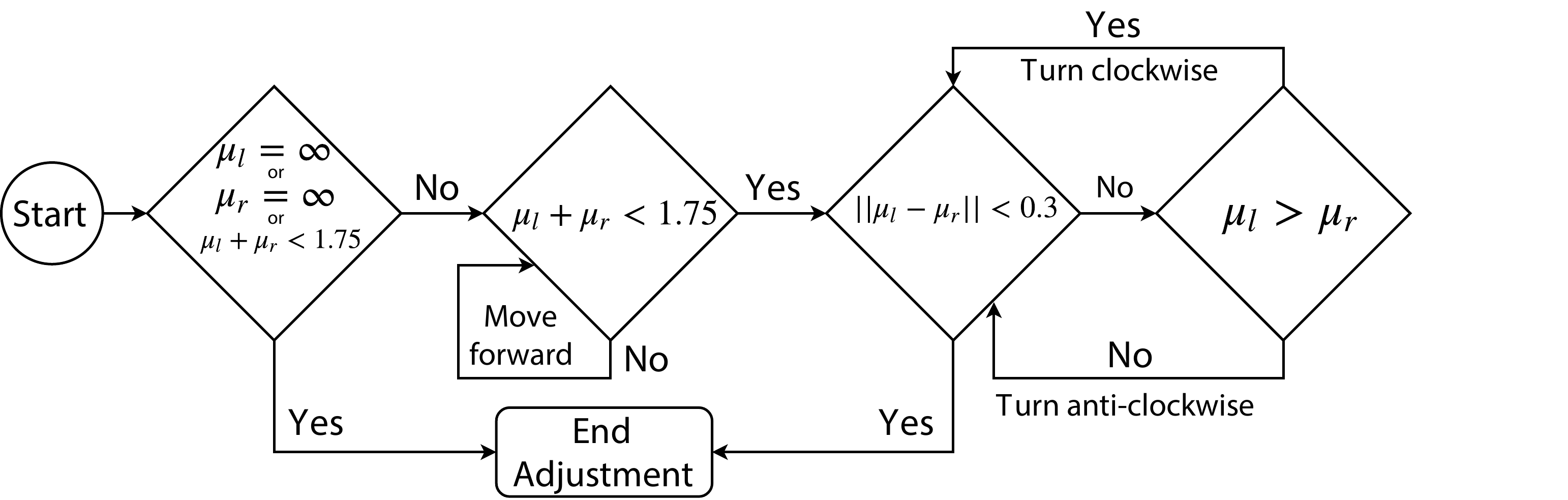}
		\caption{Process for adjusting the final pose of the Warthog.}
		\label{fig:warthog-adjust}
	\end{figure}
	
	For ensuring successful pickups, it is critical that the objects to be grasped are within the workspace of the manipulator. We accomplished this with a fine adjustment of the final base pose (Fig.~\ref{fig:warthog-adjust}) using the calculated scans from the two fixed cameras. First, the mean distances $\mu_l$ and $\mu_r$ in these scans are calculated. If one of them is infinite (no obstacles detected) or $\mu_l + \mu_r < 1.75m$, the robot does not readjust itself. Otherwise, it moves forward to decrease its distance to the objects to meet this condition. Then, if $||\mu_l - \mu_r|| < 0.3m$, no reorientation is performed. Otherwise, if $\mu_l > \mu_r$ (i.e. objects closer to the right), the robot turns clockwise to meet the previous condition and vice versa. 
	
	
	\subsection{Grasping Stack}
	\label{subsec:grasping}
	
\begin{figure}[b]
		\centering
		\includegraphics[width = 0.243\textwidth, clip = true, trim = 0 50 0 0]{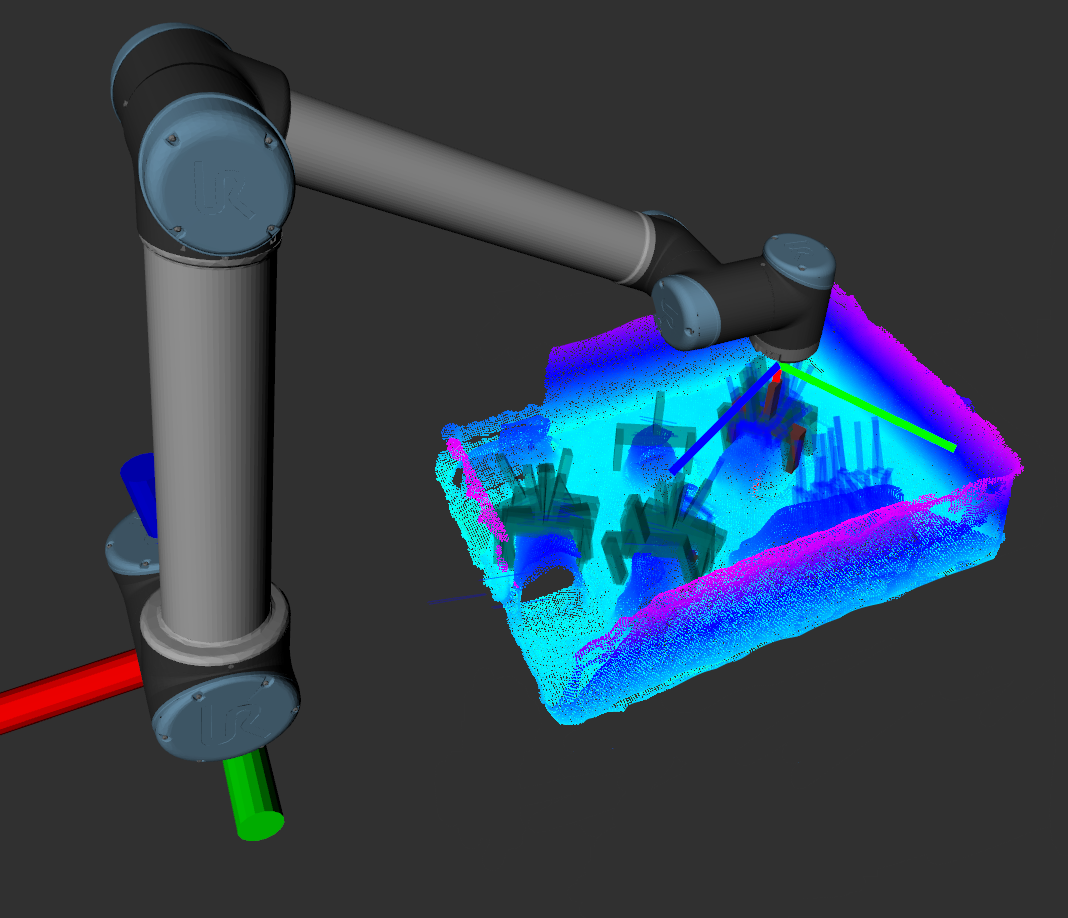}
		\includegraphics[width = 0.235\textwidth, clip = true, trim = 0 50 0 0]{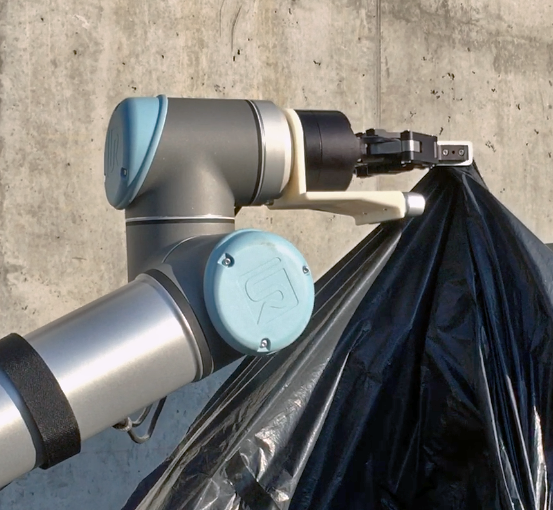}
		\caption{(left) UR10 and calculated grasps as seen in RViz while picking an object from the basket, (right) wrist pose for checking hand-eye camera.}
		\label{fig:grasping-rviz}
	\end{figure}

	We calculate grasps using the Grasp Pose Detection package (GPD) \cite{Pas2017}. This method calculates grasps poses given a 3D point cloud without having to segment the objects: the method samples grasps hypotheses over the point cloud (500 seeds in our case) and ranks their potential success using a custom grasp descriptor. Then, it returns the top K grasps found (50 in our setup). Before selecting the best grasp, we prune kinematically infeasible grasps by checking inverse kinematics (IK) solutions against the environment constraints. The system (arm, gripper, equipment on the Warthog and the Warthog itself) was modeled in OpenRave and registered point clouds were incorporated into the model. In addition, we add a flat object to act as the assumed floor at the target grasping area. Thus, using the IK solver, we discard grasp poses that are in collision with obstacles or are otherwise unreachable. The result is shown in Fig.~\ref{fig:grasping-rviz}.
	
	
	We use a set of rules to rank the remaining feasible grasps in order to find the best grasp. This rank is:
	
	\begin{equation}
		R = w h v
	\end{equation}
	
	\noindent where $w, h$ and $v$ are:
	
	\begin{align} \label{eq:grasp-selection-vals}
		w &= 1.0 - \frac{max(0.0,\gamma - min(||\theta - \alpha||, ||\theta - \beta||))}{\gamma}\\
		h &= 0.125 ||g_z - h_{min}||\\
		v &= 0.25 ||X_z||
	\end{align}
	
	\noindent $\theta$ is the grasp width, $\alpha$ and $\beta$ are the aperture limits of the gripper (0.005m and 0.085m), $\gamma$ denotes the minimum clearance (0.005m) we accept between these limits and grasp width $\theta$, $g_z$ is the z-coordinate of the translation in the grasp pose, $h_{min}$ is the support surface height (either the assumed floor or the known bottom of the basket) and $X_z$ is the z-component of the $\vec{X}$ axis of the grasp pose. 
	
	Grasps with width1 $\theta$ that meet $||\theta - \alpha|| > \gamma$ and $||\theta - \beta|| > \gamma$ maximize $w$. Hence, they are preferred because they do not force the gripper to work close to its limits. Grasps whose $g_z > h_{min}$ maximize $h$, meaning that the grasp is from a high position. As a result, the system clears piles of objects starting from the top. Finally, grasps with greater $X_z$ values maximize $v$, which is desirable in order to approach the objects perpendicularly from the top.
	
\begin{figure}[t]
		\centering
		\includegraphics[width = 0.34\textwidth]{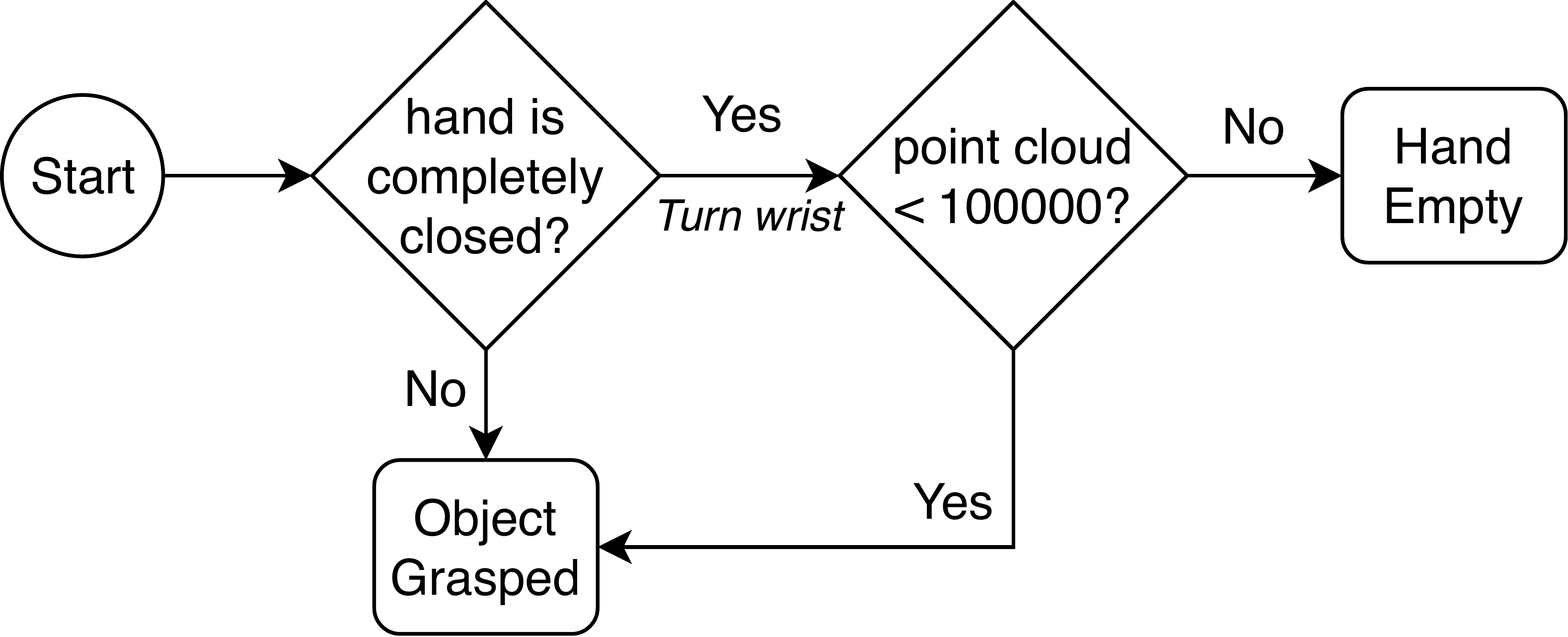}
		\caption{Process for checking the hand state after grasping.}
		\label{fig:hand-state}
	\end{figure}

	After grasping and lifting an object, we perform some tests to check for a successful pickup (Fig.~\ref{fig:hand-state}). First, we check whether the gripper is partially open (after having executed the close-gripper command to grasp). If so, we know that an object obstructs the gripper and we assume a successful grasp has occurred. If the gripper is completely closed, we must perform an additional test to check whether a thin object has been grasped. To do so, we rotate the wrist so that the hand-eye camera is below the gripper and pointing forward (Fig.~\ref{fig:grasping-rviz} right). Working on the assumption that thin objects will hang down from the grasp point (this is what trash bags do), we check whether the number of points in this point cloud is below a threshold ($100000$ in our experiments). If this condition is met, we conclude that the field of view is occluded and the pickup is successful.
	
	
	The grasp process applied for grasping from the floor or the basket is identical with the only differences being the point cloud used and the dropping place: 1) when picking an object from the floor the point cloud is acquired using the fixed cameras and the dropping point is on the basket, 2) when picking an object from the basket the point cloud is acquired with the hand-eye camera moving the arm to three view points and the drop point is in front of the robot. In order to calculate this drop point, we register a point cloud $C$ using the two fixed cameras, where $p = (p_x, p_y, p_z), p \in C$. Then, we remove the points that meet $p_z <= h_{min} + 0.05m$, where $h_{min}$ is the assumed height of the floor. The remaining are clustered and the biggest cluster $C_{bin} \subseteq C$ is assumed to be the collection bin. Then, the target position of the arm is set to a point $t = (t_x, t_y, t_z)$, where:
	
	\begin{align} \label{eq:drop-point}
		t_x &= \frac{1}{|C_{bin}|} \sum_{p \in C_{bin}}^{}p_x\\
		t_y &= \frac{1}{|C_{bin}|} \sum_{p \in C_{bin}}^{}p_y\\
		t_z &= \underset{p \in C_{bin}}{max}\{p_z\}  + 0.30
	\end{align}
	
	\noindent
	we add 0.30m to $t_z$ in order to leave some space between the drop object and the arm. In case that $|C_{bin}| < 10000$, we use a default position in front of the robot. Finally, the orientation is fixed to have the gripper pointing down.
	
	\subsection{Task Planner}
	\label{subsec:task-planner}
	
	The task planner node is in charge of sending goals to the Warthog and requests to the grasping service in order to provide the mobile manipulation functionality. It requires three inputs: the type of task, the pick position, and the drop position. Two tasks are considered:
	
	\begin{itemize}
		\item \textbf{Collect all:} the robot must collect everything from the pick point before moving to the drop point.
		\item \textbf{Collect one by one:} the robot moves between the pick and drop points transporting only one object at a time.
	\end{itemize}
	
	The type of task is passed as an argument to the task node on launch. For the pick and drop points, the RViz window from the navigation side is used as the human interface. By clicking on a position in the map using the \textit{2D Nav Goal} functionality (top bar in Fig.~\ref{fig:navigation-rviz}), the user sets goals for the task. The first set goal is the pick point and the second one is the drop point. Afterwards, the autonomous task can start:
	
	\begin{enumerate}
		\item \textbf{Moving to pick point:} the task planner sends the pick position to the Warthog, waiting for this goal to be accomplished. After reaching the pick point, the Warthog adjust its final position (see section \ref{subsec:nagivation}).
		\item \textbf{Collecting:} a request is sent to the grasping service specifying that it has to perform grasps on the floor and drops in the basket. If this is a \textit{collect all} task, this request is sent until no more grasps are found in the floor, meaning that there are no more objects left.
		\item \textbf{Moving to drop point:} the task planner sends the drop position as the new goal to the Warthog, waiting for this goal to be accomplished. Then, the Warthog adjust its final position but this time with respect to the bin.
		\item \textbf{Dropping:} a request is sent to the grasping service indicating that this time it has to perform grasps in the basket and drops in the bin. Again, if its a \textit{collect all} task this is done until no more grasps are found.
	\end{enumerate}
	
	For \textit{collect all} tasks, only a single pass through these steps is needed. For a \textit{collect one by one} task, these steps are repeated until no more objects are detected at the pick point.
	
	\section{EXPERIMENTS}
	\label{sec:experimentation}
	
\begin{figure}[t]
		\centering
		\includegraphics[width = 0.30\textwidth]{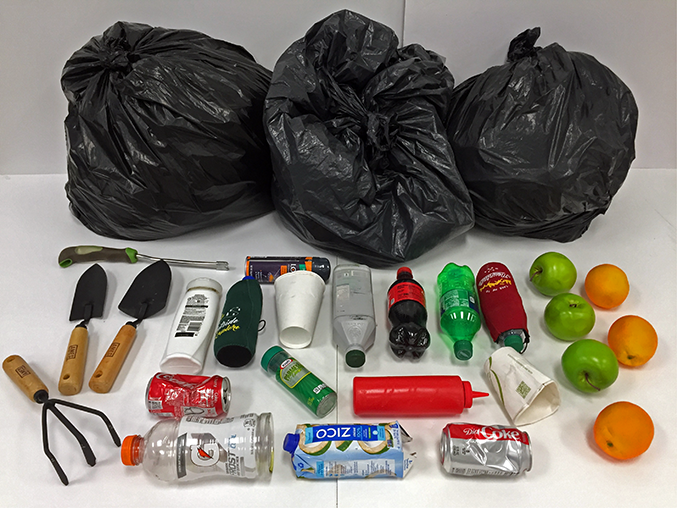}
		\caption{Test objects: trash bags, gardening tools, general garbage and fruits.}
		\label{fig:test-objects}
	\end{figure}

	We performed experiments to evaluate the variety of objects the system can handle and the success rates and times of various parts of the process. We performed the experiments on city streets in the vicinity of a loading dock as shown in Fig.~\ref{fig:env-set}. On each trial, we dropped a set of objects at a random location, placed the bin at a different random location and started the robot from a third random location. Fig~\ref{fig:test-objects} shows the set of objects used in these experiments, that were selected to be graspable by our gripper:
	
	\begin{figure}[b]
		\centering
		\includegraphics[width = 0.45\textwidth]{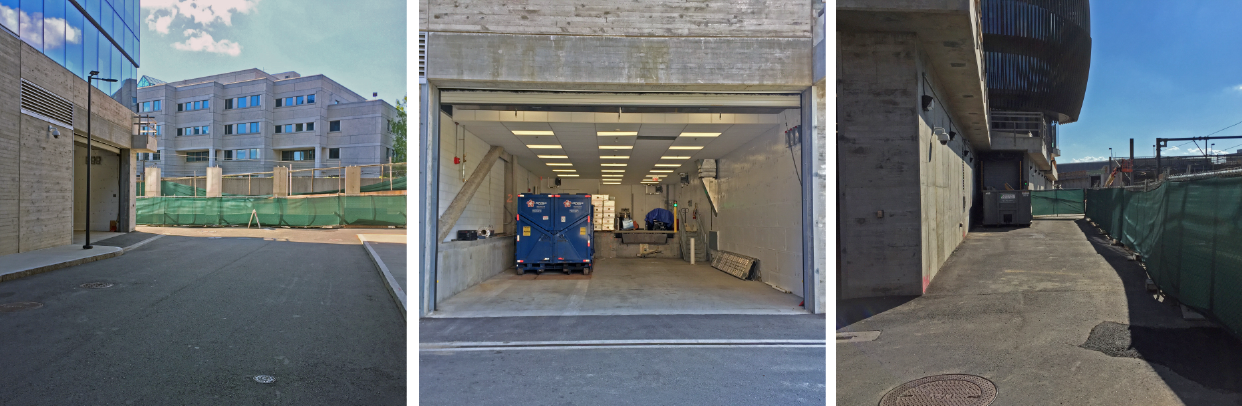}
		\caption{Testing area: main street, loading dock and narrow street.}
		\label{fig:env-set}
	\end{figure}
	
	\begin{itemize}
		\item \textbf{Trash bags:} 3 black trash bags made of plastic, which are deformable so their shape changed from test to test.
		
		\item \textbf{General garbage:} 15 objects that could be found lying in the street like plastic bottles, cans and paper cups.
		
		\item \textbf{Gardening tools:} 4 gardening tools made of steel with wood handles, except for one with rubber handle.
		
		\item \textbf{Fruits:} 3 green apples and 3 oranges.
	\end{itemize}
	
	
	The navigation subsystem was evaluated in terms of the number of plans needed to move from one point to another (e.g. going from the pick to the drop point). If just one plan was needed, that was a 100\% success rate. If in the way the robot got lost or stuck, the \textit{move\_base} package stopped the robot. Thus, a new plan was needed to reach the goal from the current position. If that second attempt was successful, the success rate was 50\% because two plans were required. The grasping subsystem was evaluated in terms of the grasp success rate. A grasp was considered to be successful only if the desired object was grasped and deposited in the bucket or the bin. The end-to-end task was considered a success only if all of the items were transported from the pick point to the place point without human intervention.
	
\begin{figure}[b]
		\centering
		\includegraphics[width = 0.43\textwidth]{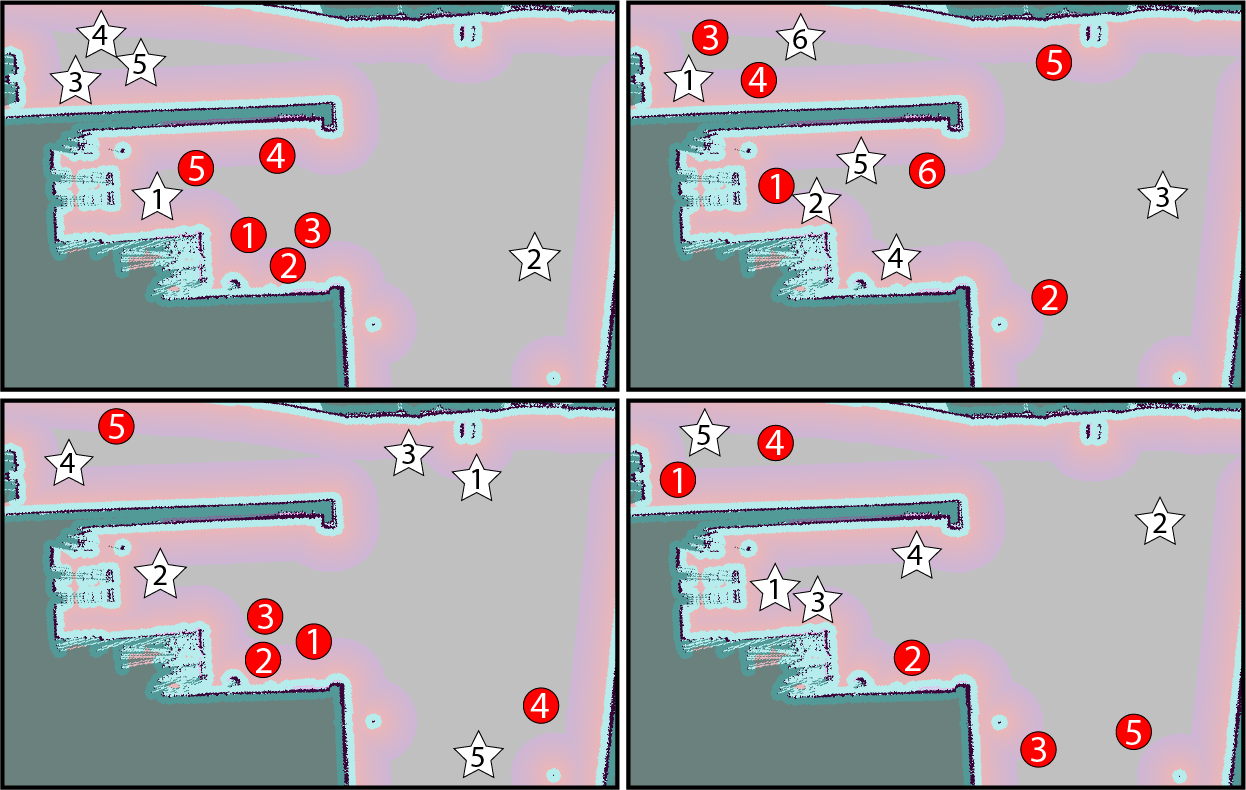}
		\caption{Maps showing pick (stars) and drop (circles) points for each experiment: (top-left) trash bags, (top-right) general garbage, (bottom-left) gardening tools and (bottom-right) fruits. Numbers indicate the trial pairs.}
		\label{fig:tasks-maps}
	\end{figure}

\begin{figure*}[ht]
		\centering
		\includegraphics[width = \textwidth]{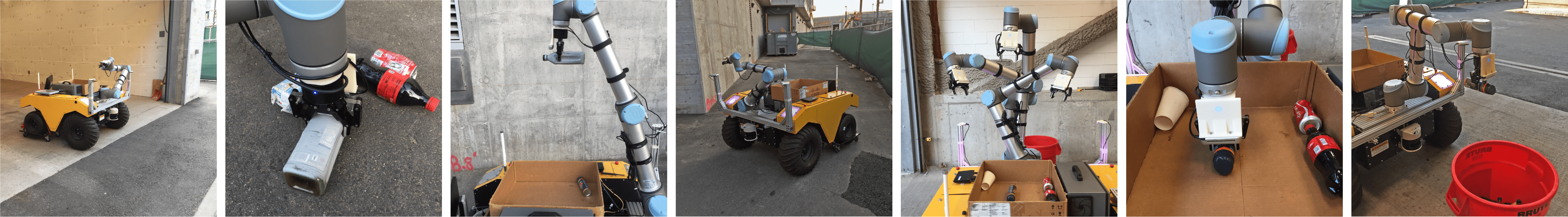}
		\caption{Sequence of actions taken by the robot during a trial: moving towards the pick point, grasping an object from the floor, dropping it in the basket, moving to drop point, registering three views from the basket, grasping an object from it and finally dropping the object in the collecting bin.}
		\label{fig:trial}
		\vspace{-4mm}
	\end{figure*}

	In total, we performed four experiments, one for each of the objects we considered: trash bags, general garbage, gardening tools and fruits. In each of the four scenarios, we ran between five and six task trials. The randomly generated pick and drop points for each scenario are shown in the generated map in Fig.~\ref{fig:tasks-maps}. Results are summarized in Table \ref{table:results} and Table \ref{table:times}. Fig.~\ref{fig:trial} shows one sequence for one trial.
	
	\subsection{Trash Bags}
	
	We performed five trials for the trash bag scenario. Since one trash bag was big enough to fill the basket, these tests were executed following the \textit{one by one} method. The navigation success rate was 92.1\%. In 3 occasions the robot needed a second attempt, mainly when it found itself too close to an obstacle, like when moving away from a narrow space. The grasping success rate from the floor was 78.9\%. The robot performed 3 grasps that did not grip the object and in 1 case the bag slipped from the gripper while lifting it. The success rate when grasping from the basket was 100.0\%.
	
	\subsection{General Garbage}
	
	\begin{table}[t]
		\centering
		\caption{Achieved success rates on each set. \textit{P/D-nav} is the task of moving to pick/drop points and \textit{P/D-grasp} is grasping at them.}
		\label{table:results}
		\begin{tabular}{@{}cccccc@{}}
			\toprule
			\textbf{Set -- trials} & \textbf{P-Nav} & \textbf{P-Grasp} & \textbf{D-Nav} & \textbf{D-Grasp} & \textbf{Task} \\ \midrule
			\rowcolor[HTML]{EFEFEF} 
			Bags -- 5 & \begin{tabular}[c]{@{}c@{}}20/23\\ (87\%)\end{tabular} & \begin{tabular}[c]{@{}c@{}}15/19\\ (79\%)\end{tabular} & \begin{tabular}[c]{@{}c@{}}15/15\\ (100\%)\end{tabular} & \begin{tabular}[c]{@{}c@{}}15/15\\ (100\%)\end{tabular} & \begin{tabular}[c]{@{}c@{}}5/5\\ (100\%)\end{tabular} \\
			Garbage -- 6 & \begin{tabular}[c]{@{}c@{}}6/6\\ (100\%)\end{tabular} & \begin{tabular}[c]{@{}c@{}}50/56\\ (89\%)\end{tabular} & \begin{tabular}[c]{@{}c@{}}6/8\\ (75\%)\end{tabular} & \begin{tabular}[c]{@{}c@{}}50/60\\ (83\%)\end{tabular} & \begin{tabular}[c]{@{}c@{}}5/6\\ (83\%)\end{tabular} \\
			\rowcolor[HTML]{EFEFEF} 
			Tools -- 5 & \begin{tabular}[c]{@{}c@{}}5/5\\ (100\%)\end{tabular} & \begin{tabular}[c]{@{}c@{}}17/28\\ (61\%)\end{tabular} & \begin{tabular}[c]{@{}c@{}}5/5\\ (100\%)\end{tabular} & \begin{tabular}[c]{@{}c@{}}18/27\\ (67\%)\end{tabular} & \begin{tabular}[c]{@{}c@{}}3/5\\ (60\%)\end{tabular} \\
			Fruits -- 5 & \begin{tabular}[c]{@{}c@{}}5/5\\ (100\%)\end{tabular} & \begin{tabular}[c]{@{}c@{}}30/33\\ (91\%)\end{tabular} & \begin{tabular}[c]{@{}c@{}}5/5\\ (100\%)\end{tabular} & \begin{tabular}[c]{@{}c@{}}30/39\\ (77\%)\end{tabular} & \begin{tabular}[c]{@{}c@{}}5/5\\ (100\%)\end{tabular} \\ \bottomrule
		\end{tabular}
	\end{table}

	We performed six trials for the garbage collection scenario. In this case, every trial followed the \textit{collect all} method. Since we had 15 test objects, we randomly sampled seven on each trial except for one in which the whole set was used. The robot achieved 85.7\% success rate on navigation. One navigation needed a second plan while moving away from a wall. The other failure was caused by the drift of the IMU integrated in the robot: it made the system represent behind the robot a near wall in the map. Only in this occasion we manually turned the robot to update the local map. After that, the robot moved autonomously to the other point.
	
	
	Grasping general garbage from the floor was more challenging: the system achieved a 89.3\% success rate. Since these objects are smaller, their point clouds are less accurate and grasps need to be more precise as well. Out of the 6 failures, 3 were caused by the wind moving an object during the point cloud registration. There was also 1 failure caused by a poor grip that did not contact the object, 1 slip while lifting the object and 1 reattempt because the planner could not find a collision free trajectory for reaching the best grasp.
	
	Finally, grasping these objects from the basket yielded an 83.3\% success rate. From the 10 failures, 4 were slips while lifting the object, mainly because the objects were in the corners of the basket, making it difficult for the GPD to find good grasps. Then, 3 failures were caused by the wind moving the objects while registering the point cloud. The last 3 failures were grasps that did not grip the object correctly.
	
	\subsection{Gardening Tools}
	
	In this experiment five trials were executed following the \textit{collect all} method. In 4 out of 5 navigations to the pick point the final readjustment of the base failed ($\mu_r$ or $\mu_l$ were infinite). However, we report these navigations as successes because the system reached the user pick point with just one plan. The fact that grasps at the pick point were executed from the user estimated position probably contributes to the lower grasping success rates achieved with these objects.
	
	
	The gardening tools were the most challenging to pick up from the ground. In one trial the robot could not detect one of the tools properly so it moved to the drop point leaving it behind at the pick point. This happened again in another trial, in which two objects were left behind. In consequence, the success rate was 60.7\%. From the 11 failures, 10 of them were caused by the gripper performing a weak power grasp. Since the handles are thin, sometimes the gripper closed around them but leaving enough room for them to slip. The other failure was an attempt to grasp the tines of the rake.
	
	When the robot left behind an object, we poured it in the basket manually in a random position for testing the performance when grasping the tools from it. The success rate for grasps from the basket was 66.7\%. In 2 occasions the object fell out of the bin when dropping it because it was long and most of its volume was out of the bin while falling. There were 2 slips when lifting and 2 grasps that did not grip the target. Finally, in 2 more cases the grasp failed because the robot tried to grasp the tines of the rake.
	
	\subsection{Fruits}
	
	Finally, five trials were executed in this experiment, following the \textit{collect all} method. The navigation success rate was 100.0\%. The grasping success rate for picking fruits from the floor was 90.9\%. There were 3 failures caused by two objects being so close together that the robot attempted to grasp them at the same time from their contacting side. When grasping the fruits from the basket, the system achieved a 76.9\% success rate. From the 9 failures, 3 were caused by poor grasps that did not grip the object enough. In 2 cases, the object slipped while lifting it. The remaining 4 failures were grasping attempts performed over artifacts registered in the cloud due to direct sunlight in the camera.
	
	\subsection{Execution Time}
	
	The time required for moving from one point to another principally depended on the distance between the points and the velocity of the robot (0.5m/s in our tests). In general, moving an object between two points took 118s in average (44s minimum; 239s maximum). The time required to register the point cloud was low-variance, but differed depending upon whether it was performed using the two fixed cameras or the hand-eye camera. With the two fixed cameras, registration took 4s on average. When it was performed using the hand-eye camera, it took 33s in average since it had to move the arm to three views. Calculating grasps required 10s on average for the cloud registered with the fixed cameras and 18s for the one stitched with the hand-eye camera. It took more time to process the hand-eye cloud because it uses three views and therefore contains more points. Finally, pick execution and drop execution took similar amounts of time: 48s in average for grasps from the ground and 51s from the basket.
	
	\begin{table}[t]
		\centering
		\caption{Execution time of each process of the transportation task.}
		\label{table:times}
		\begin{tabular}{@{}lcc@{}}
			\toprule
			\multicolumn{1}{c}{\textbf{Sub-process}} & \textbf{Pick} & \textbf{Drop} \\ \midrule
			Register Point Cloud & 3.78s $\pm$ 0.21s & 33.20s $\pm$ 9.00s \\
			Calculate Grasp & 10.14s $\pm$ 5.41s & 17.72s $\pm$ 8.05s \\
			Execute Grasp & 48.12s $\pm$ 7.09s & 50.86s $\pm$ 13.72s \\
			Navigate to Point & 98.18s $\pm$ 25.73s & 137.86s $\pm$ 39.49s \\ \bottomrule
		\end{tabular}
	\end{table}
	
	
	\section{CONCLUSIONS AND LIMITATIONS}
	\label{sec:conclusions}
	
	This paper describes a system that solves an open world transportation task involving novel objects. After being provided with a pick and a drop point by a user, our system autonomously navigates to the pickup point, grasps everything there, navigates to the dropoff point, and drops everything into a bin. We evaluated the system in four experimental scenarios involving the following different objects: trash bags, garbage, tools and fruits. The experiments indicate that our system worked relatively well, yielding an 80.8\% grasping success rate, navigating without problems 96.1\% of the cases, and giving an 85.7\% overall task success rate.

	
	However, the system has some limitations. Since it uses a 2D laser scanner, it has problems localizing itself in areas with elevation changes. This confuses the system so that the robot oscillates while traversing them. In experimentation, we could not set goals in the entrance of the loading dock because that area was really depressed compared to the rest. As for the grasping system, it has difficulties to grasp objects that do not rise from the ground more than 3cm, approximately. The D415 cameras record noise when working outdoors that increases with the distance, so these objects are hard to distinguish in the floor and could be left behind undetected, like happened with the gardening tools.
	
	As a future work, we want to reduce the time gap from registering the cloud to actually performing a grasp. Since the robot works in an open environment, there are factors that can affect the position of the objects, like we experienced with the wind moving them. Moreover, it would be good to include a 3D sensor to improve the self-localization. Finally, we would like to work on an object detection and tracking system so the robot can find the target objects autonomously.
	
	
	\bibliographystyle{IEEEtran}
	\bibliography{biblio}
	

	\appendices
	
	\section{Warthog Steering} \label{app:warthog-steering}
	
	The original vehicle is setup as a skid steering system with a motor on each side of the robot rigidly coupled to a gearbox that drives both the axles, allowing it to turn in place. During our initial experiments, we discovered that the skid steering coupled with the large amphibious tires added a high level of non-linearity to the system. Thus, when turning in place or with very low forward velocity, while changing rotational direction, the motors ran into their torque limits trying to overcome the potential energy stored in the deformed tires. This non-linearity implied that we could not use a standard off-the-shelf controller. 
	
	In order to keep the control of the robot simple, we disengaged the wheels on the rear axle from the gearbox and put them on vehicle skates, converting the robot to a differential drive. The reconfiguration to a differential drive system helped significantly by bringing down the time required to change rotation direction from more than 5 secs to less than 1 sec. However, this conversion had two limitations: we increased our turning radius by 0.43m and we added a non-linearity of a different form (increased non-holonomity). The vehicle skates use offset caster wheels which results in the rear of the vehicle continuing to move in the direction of the casters for a short period of time while they re-align.

\end{document}